\documentclass[letterpaper, 10 pt, conference]{ieeeconf}  
\usepackage{amsmath}
\usepackage{cite}
\usepackage{color}
\usepackage{graphicx}
\usepackage{booktabs}
\usepackage{threeparttable}
\IEEEoverridecommandlockouts                         
\title{\LARGE \bf
LESS-Map: Lightweight and Evolving Semantic Map in Parking Lots for Long-term Self-Localization
}

\author{Mingrui Liu$^{1}$, Xinyang Tang$^{2}$, Yeqiang Qian$^{2}$, Jiming Chen$^{1}$, and Liang Li$^{1}$
\thanks{$^{1}$College of Control Science and Engineering,
        Zhejiang University, Hangzhou, 310027, P. R. China.}%
\thanks{$^{2}$Department of Automation, Shanghai Jiao Tong University, Shanghai, 200240, P. R. China}%
}

\begin{document}

\maketitle
\thispagestyle{empty}
\pagestyle{empty}

\begin{abstract}
Precise and long-term stable localization is essential in parking lots for tasks like autonomous driving or autonomous valet parking, \textit{etc}. Existing methods rely on a fixed and memory-inefficient map, which lacks robust data association approaches. And it is not suitable for precise localization or long-term map maintenance. In this paper, we propose a novel mapping, localization, and map update system based on ground semantic features, utilizing low-cost cameras. We present a precise and lightweight parameterization method to establish improved data association and achieve accurate localization at centimeter-level. Furthermore, we propose a novel map update approach by implementing high-quality data association for parameterized semantic features, allowing continuous map update and refinement during re-localization, while maintaining centimeter-level accuracy. We validate the performance of the proposed method in real-world experiments and compare it against state-of-the-art algorithms. The proposed method achieves an average accuracy improvement of 5cm during the registration process. The generated maps consume only a compact size of 450 KB/km and remain adaptable to evolving environments through continuous update.

\end{abstract}

\section{Introduction}

Accurate mapping and localization are crucial for intelligent vehicles in lots of tasks. Commercial intelligent vehicles, equipped with multiple sensors, require a mapping solution that balances accuracy, efficiency, robustness, and long-term upgradability. Accuracy ensures high-quality performance, efficiency reduces processing costs, robustness enhances adaptability, and ongoing maintenance enables consistent map updates for evolving environments. More recently, visual semantic features \cite{mark} have gained much attention for localization due to their stability over time and resilience to lighting changes. Moreover, the extraction of semantic features relies solely on standard cameras available at conventional prices, which makes semantic-based mapping and localization solutions suitable for widespread use in commercial intelligent vehicles\cite{roadmap}.

\begin{figure}[t]
    \centering
    \includegraphics[width=1.0\linewidth]{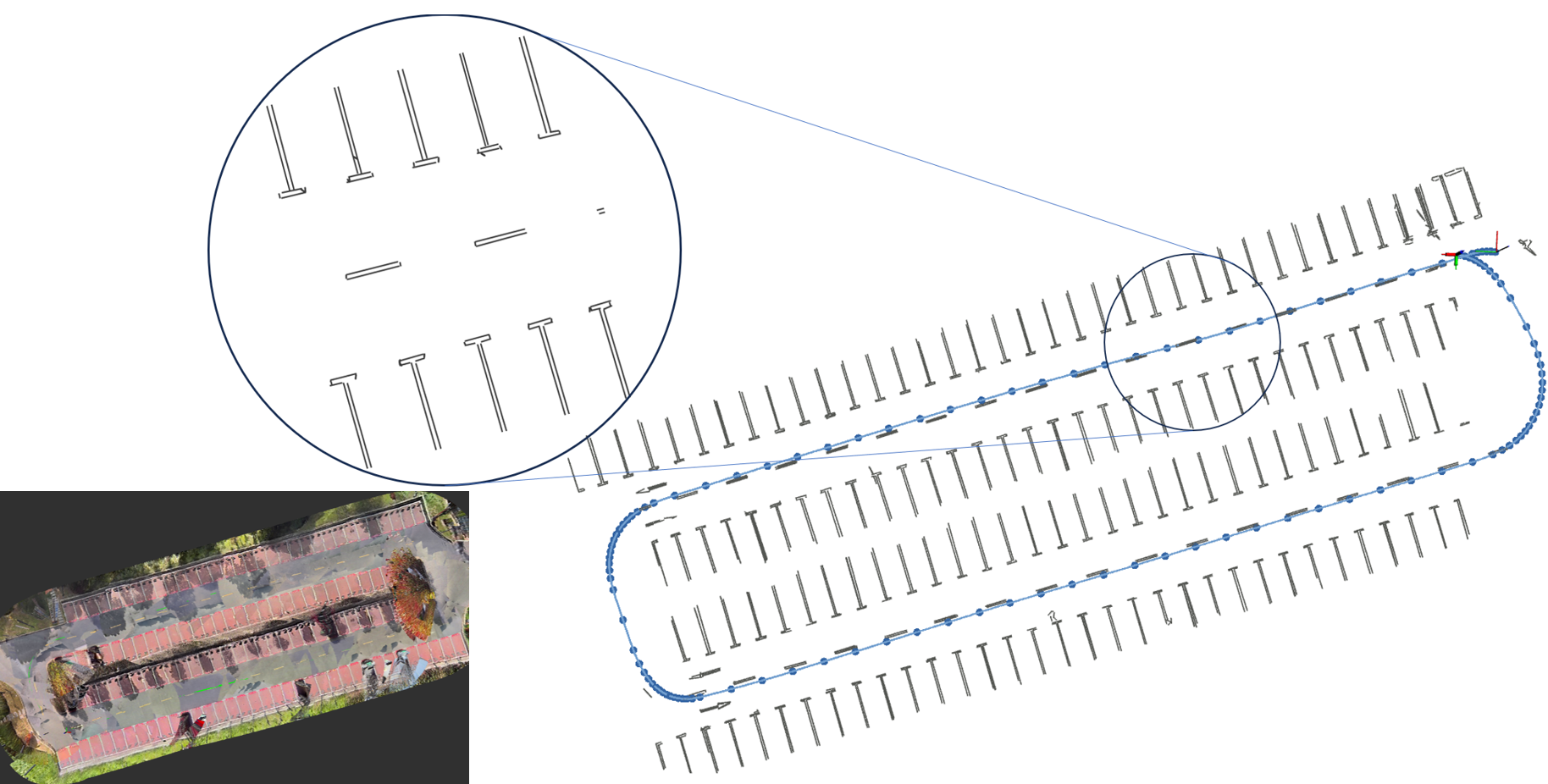}
    \caption{The figure in the left corner depicts a parking lot scene, while the larger figure represents the map we build for the parking lot. We propose a lightweight representation for semantic features, enabling centimeter-level localization accuracy for vehicles. }
    \label{fig-intro}
\end{figure}

Parking lots often contain numerous ground markers such as arrows, parking lines, stop lines, \textit{etc}\cite{avpslam}. As shown in Fig. \ref{fig-intro}, these markers have regular, flat geometric structures and are ubiquitous in fully structured parking lots. Therefore, investigating their utility in mapping and localization makes sense. Unlike traditional 3D geometric features such as planes or edges\cite{loam,liosam}, ground semantic features are solid figures firmly attached to the ground plane, with their primary geometric features embedded in their contours. A parameterized method leveraging this property can effectively facilitate high-quality data association for ground semantic features, thereby playing a positive role in registration, mapping, and localization processes. 

However, existing methods lack robust parameterization of ground semantic information, resulting in a lack of robust data association during mapping and localization processes\cite{roadslam,avpslam,fuseAVP,semICP}. This drawback adversely affects the accuracy of localization and poses limitations on the long-term update capability of the map. Hence, in this work, we present an accurate and lightweight parameterization method for ground semantic features to assist in achieving higher accuracy and efficiency in mapping and localization. Furthermore, we propose a novel map update approach, allowing continuous map update and refinement during re-localization, while maintaining centimeter-level accuracy.
The contributions of this work are summarized as follows:
\begin{itemize}
    \item We present a novel parameterization method for ground semantic features to achieve more efficient pose estimation and more robust data association.

    \item We propose a lightweight map maintenance method that can be updated by the localization results of subsequent vehicles, enabling centimeter-level localization accuracy.

    \item We provide a complete and practical pipeline of mapping, localization, and map update, demonstrated to perform reliably in both outdoor and indoor scenarios through real-world experiments.
\end{itemize}
\section{Related Works}
 \begin{figure*}[t]
    \centering
    \includegraphics[width=1.0\linewidth]{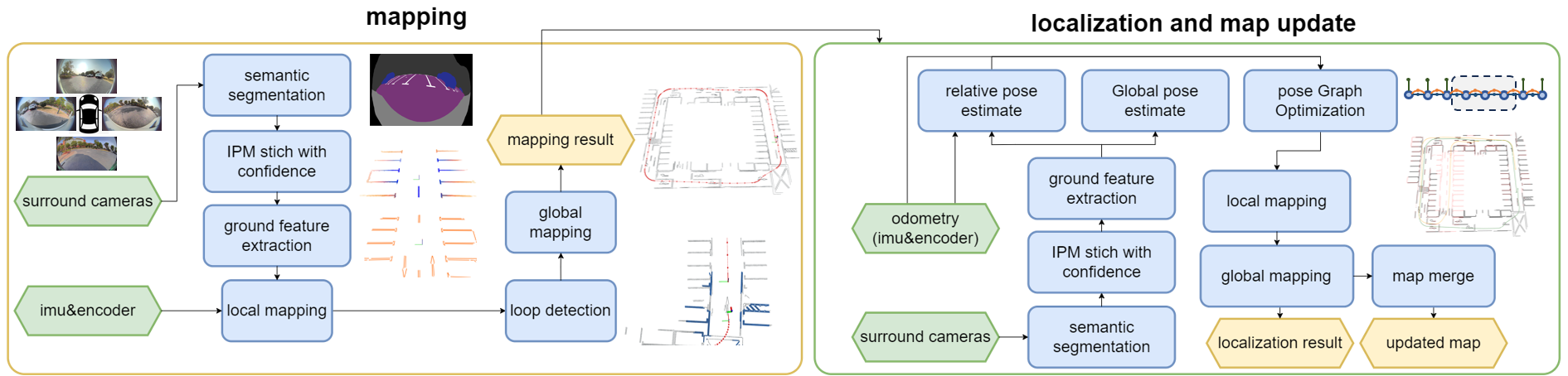}
    \caption{Overview of the proposed algorithm. We utilize four surround cameras to extract accurate parameterized ground features. The system is mainly divided into two parts. The first (left) part is the mapping part, it creates a global map when the environment is visited for the first time. The second (right) part is the localization and map update part, it engages localization and updates the prior map with newly captured environmental data.}
    \label{fig-pipeline}
\end{figure*}

In recent decades, mapping and localization in artificial environments like parking lots have gained increasing attention. LiDAR and vision-based approaches are two common and widely used solutions for accurate mapping and localization. LiDAR-based solutions\cite{loam,liosam} achieve accurate relative pose estimation by extracting line and plane features from LiDAR point cloud. To achieve a more lightweight and efficient representation of geometric features, some methods use probabilistic models such as the Gaussian mixture model\cite{gmmMap}, normal distribution model\cite{voxelMap}, and Gaussian process\cite{gpMap} for mapping. These approaches can also be applied to parameterize artificial environments\cite{parkingStructual}. 
Nevertheless, in parking lot scenarios, LiDAR-based methods need to deal with the influence of semi-static and dynamic vehicles in the scene. Additionally, the high cost of LiDAR systems makes them unsuitable for widespread deployment.

Vision-based methods have gained popularity due to the low cost of cameras. Traditional methods extract visual keypoints and descriptors for scene description\cite{orbslam2,vio,monovio,orbslam3}, but they suffer from large map storage, sensitivity to dynamic objects, and illumination variations. Some methods focus on extracting geometric features such as lines\cite{lineSLAM,lineStructural}, planes\cite{planeSLAM}, and cubes\cite{cubeslam}. These methods avoid the vulnerability of pixel-level features and are more robust in artificial environments, but they struggle with stable feature extraction in parking lots. Some methods extract highly defined features like parking slots\cite{parkingslot,slotdetect} for localization. However, fluctuating perception results of these methods will greatly affect the accuracy. Other methods extract ground marks in parking lots \cite{roadslam,fuseAVP,semICP} to build the localization-oriented map since ground marks can be stably observed in most cases. Many works utilize neural networks to obtain semantic information for mapping and localization. For instance, \cite{avpslam} adopts a semantic segmentation model to extract ground semantic information such as guide signs and parking lines in the underground parking lots, and an ICP method is used to estimate the vehicle's pose. \cite{roadmap} integrates ground semantic features into grid maps for mapping and employs a voting-based map updating strategy. However, these methods employ simple merging techniques for maps and do not adequately describe the geometric structure of ground features. As a result, the accuracy of localization is compromised.
In this work, we focus on using ground semantic features to build a complete system for mapping, localization, and map update. Our approach emphasizes the generation of a precise yet lightweight map representation, enabling stable map updates and long-term localization.

\section{Methodology}

\subsection{System Overview}

We utilize four surround-view fisheye cameras, an Inertial Measurement Unit (IMU), and wheel encoders to capture the surroundings and estimate the vehicle's trajectory. The proposed algorithm comprises two main components: mapping and localization, as illustrated in Fig. \ref{fig-pipeline}.

\textbf{Mapping:} When a vehicle enters the environment for the first time, a fundamental task is to construct a global map as the foundation for subsequent localization and map update. We extract ground semantic features from surround-view cameras and project them into the 3D space. The uncertainties of these observations are estimated to enhance the map fusion. These features are then parameterized using the method detailed in Section \ref{parameterization}. As a result, the global map is generated through a combination of pose estimation and refinement of loop closures.

\textbf{Localization and Map Update:}
After constructing the global map, upon the vehicle's return to the environment, it can engage in localization and map update using newly captured environmental data. 
In light of the dynamic nature of the environment and the possibility of going into unmapped areas, a factor graph is constructed to fuse odometry and localization results, yielding a fine-tuned global trajectory and a newly generated map. The newly generated map is then seamlessly merged with the existing map, ensuring a robust and consistent map update process.


\subsection{Semantic Segmentation and Cloud Projection}
\label{sectionSemantic}

We employ a robust segmentation network\cite{segnet} to effectively detect ground surfaces, lane lines, and road markers from the original fisheye images. 
After semantic segmentation, we transform the fisheye images into a semantic point cloud within the vehicle coordinate system. This transformation involves undistorting the fisheye images and performing an inverse perspective mapping (IPM)\cite{adaptiveIPM}. 

 \begin{figure}[h]
    \centering
    \includegraphics[width=0.9\linewidth]{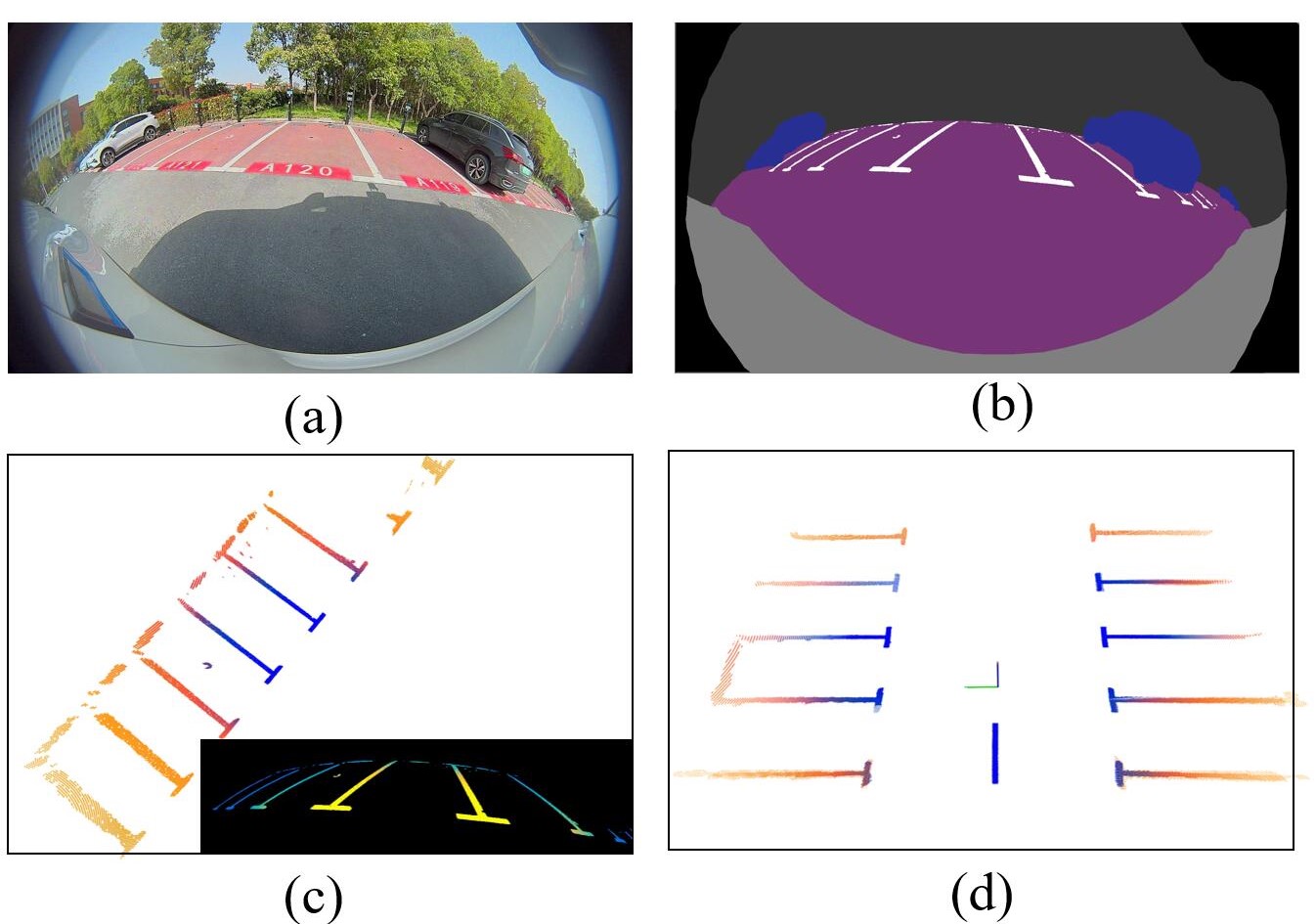}
    \caption{Cloud Projection results. (a) Raw image captured by a fisheye camera. (b) Result of the semantic segmentation. (c) Point cloud obtained by IPM projection, rendered with confidence measures. (d) A complete point cloud stitched by the results of four fisheye cameras.
}
    \label{fig-segmentation}
\end{figure}

The fisheye camera is modeled using a practical fisheye camera model \cite{scaramuzza}, which represents the camera's projection model as follows: 
\begin{equation}
\begin{aligned} \begin{bmatrix} X_c \\Y_c\\Z_c \end{bmatrix} &=\lambda \begin{bmatrix} u_c\\v_c\\ d(\rho) \end{bmatrix}, \rho=\sqrt{u_c^2+v_c^2} \\ \begin{bmatrix} u\\v \end{bmatrix} &=S_{2\times2} \begin{bmatrix} u_c\\v_c \end{bmatrix} + \begin{bmatrix} u_0\\v_0 \end{bmatrix}  \end{aligned}
\end{equation}
where $u$ and $v$ are pixel coordinates in the image, $u_c$ and $v_c$ are pixel coordinates with the camera optical axis as the origin, $u_0$ and $v_0$ are the coordinates of the optical axis center in the pixel coordinate system, $\lambda$ is a scale factor, $d(\rho)$ represents the distortion polynomial $d(\rho)=a_0+a_2\rho^2+a_3\rho^3+a_4\rho^4$, and $S$ is the stretch matrix. Thus, the undistorted coordinates $u',v'$ corresponding to the pixel coordinates $u,v$ in the fisheye image can be expressed as:
\begin{equation}
\begin{aligned} \begin{bmatrix} u'\\v'\end{bmatrix}= \begin{bmatrix}  \frac{a_0}{d(\rho)}u+(1-\frac{a_0}{d(\rho)})u_c\\ \frac{a_0}{d(\rho)}v+(1-\frac{a_0}{d(\rho)})v_c\\ \end{bmatrix}
\end{aligned}
\end{equation}
After obtaining the undistorted pixel coordinates $u',v'$, the point cloud $(x_b,y_b)$ in the vehicle coordinate system can be obtained using a pre-calibrated inverse perspective matrix $H$: 
\begin{equation}
 \mu\begin{bmatrix}x_b\\y_b\\1\end{bmatrix} =H_{3\times3}^{-1}\begin{bmatrix}u'\\v'\\1\end{bmatrix}
\end{equation}

To facilitate subsequent local map fusion, we calculate a confidence measure during the projection process, which serves as an evaluation of the observation quality for the projected semantic point cloud. Based on the above calculations, the distance $\sqrt{x_b^2+y_b^2}$ of $(x_b,y_b)$ from the vehicle center can be expressed as a function $f(u,v)$ of the pixel coordinates $(u,v)$. The gradient $||\nabla f||$ of this function represents the size of the pixel in physical space. A larger $||\nabla f||$ indicates greater uncertainty of the pixel in physical space. Since the magnitude of $||\nabla f||$ is related to the camera model itself, we define a confidence measure for each pixel using a normalization function: 
\begin{equation}
bel([u,v])=\frac{1}{1+\exp(b(a-1/||\nabla f||))}
\end{equation}
where $a$ and $b$ are two parameters related to factors such as image size, distortion model, and camera installation position. This confidence measure represents the observation quality of the camera projection model within a specific region. The visualization of this confidence measure is shown in Fig. \ref{fig-segmentation}. Notably, since the mapping for projection remains constant for a given camera, the above steps only need to be computed once during camera initialization and do not require repetitive computation.
 
\subsection{Contour Parameterization}
\label{parameterization}
\begin{figure}[b]
    \centering
    \includegraphics[width=0.9\linewidth]{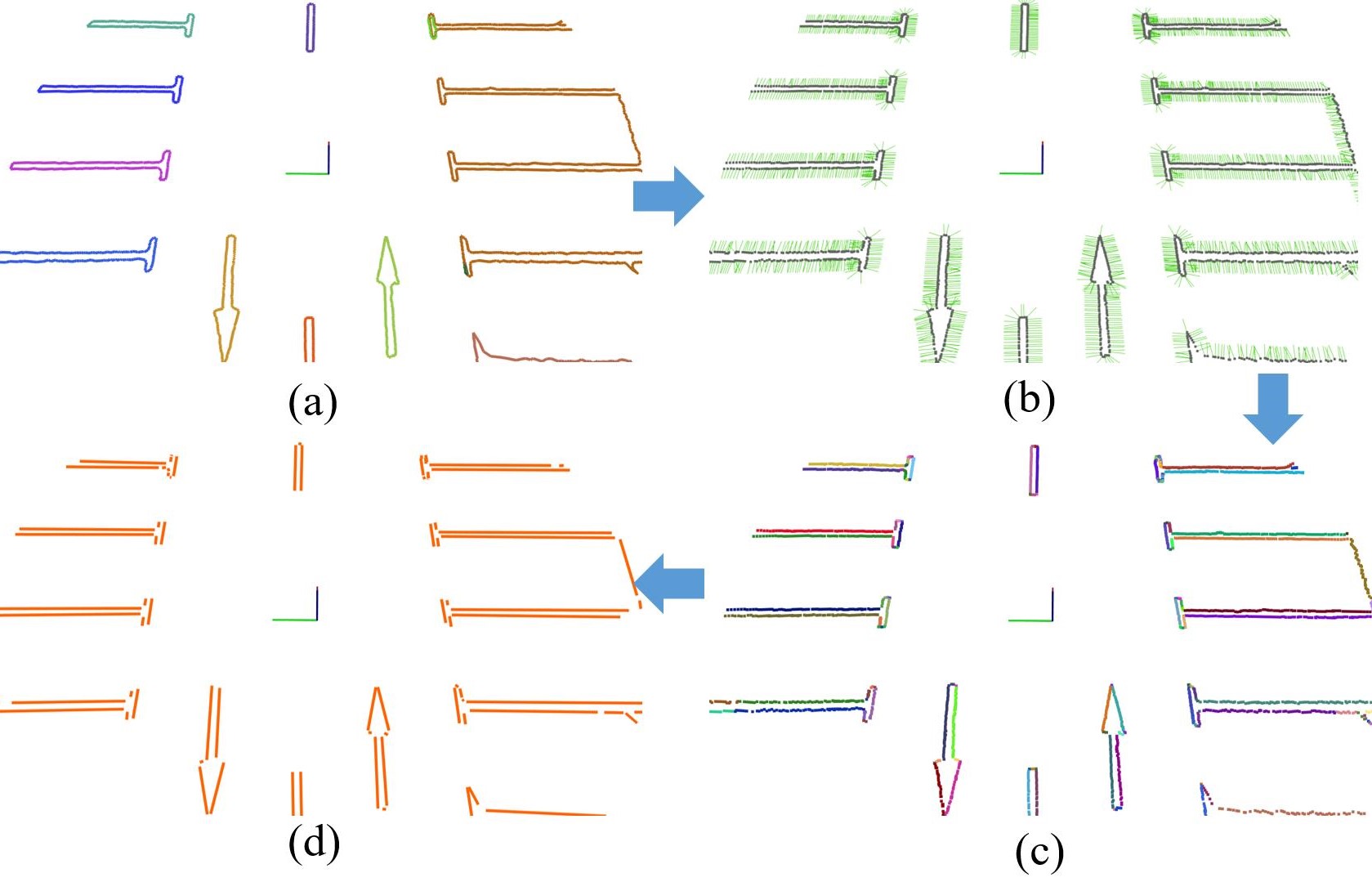}
    \caption{Our parameterization process for ground semantic features. (a) Different contours in the point cloud.  (b) Contour normal vectors (green lines) of contour points. (c) Line clusters obtained by region-growth algorithm. (d) Line features extracted from the line cluster cloud. }
    \label{fig-parameterize}
\end{figure}

In order to fully leverage the semantic structure of the ground, we propose an adaptive parameterization approach to encode ground semantic information. The step-by-step process of this method is illustrated in Fig. \ref{fig-parameterize}.

We employ the border-following method to extract the contours of road markers in the image with semantic labels. 
Utilizing the approach detailed in Section \ref{sectionSemantic}, we obtain the contour point cloud $P_b^t$ in the vehicle coordinate system. Furthermore, the adjacency relationship, label (index) of each contour, and confidence of each point are also stored in $P^t_b$.  In order to eliminate trivial or unreliable features that may exist in noisy environments, contours with low confidence and small areas are filtered out.  The contour-clustering result is shown in Fig. \ref{fig-parameterize}(a).

After generating $P_b^t$, we calculate the 2D contour normal vector for each contour point since the point clouds obtained through IPM projection all lie on the same plane. This contour normal vector is defined as the vector pointing from the contour's interior to its exterior, perpendicular to the contour's edge. We utilize a surface-based edge estimation algorithm \cite{contour} to estimate the contour normal vectors. The estimated result is shown in Fig. \ref{fig-parameterize}(b).
Then, according to the adjacency relationships we obtained, we employ a region-growth algorithm to cluster the contour points after obtaining the contour normal vectors.
Points with consistent contour normal vectors are clustered into the same group. The clustering result is shown in Fig. \ref{fig-parameterize}(c), illustrating a process that splits a road marker into multiple line clouds. 
Subsequently, for each line cluster, we calculate its direction and centroids to obtain the 6-DOF line coefficients. In addition, a set of endpoints for each line cluster is also obtained. Extracted line features are shown in Fig. \ref{fig-parameterize}(d). We use the average confidence measure of all points in a line cluster to represent the confidence of the line.

In summary, we regard a frame of the raw point cloud as a set of lines denoted as $L_t=\{l_1,l_2,...,l_k\}$. Each line is characterized by a 6-DOF coefficient $coef_i$, a contour normal vector $\boldsymbol{e}_{i}$, two endpoints $p_{d1}$ and $p_{d2}$, a cluster label $k_i$, and an observation confidence $c_i$.

\subsection{Odometry}
\label{odom}

The odometry estimates the pose transformation between two successive camera shots. Building upon the features we extract in Section \ref{parameterization}, we utilize $P_b^t, L_t$, and $P_b^{t-1}, L_{t-1}$ to estimate the relative pose transformation $^{t-1}_tT$. This estimation is carried out via a point-to-line ICP method, with several modifications implemented to improve both accuracy and efficiency.

\textbf{Contour normal vector assisted point-line matching.} We use contour normal vectors to find better correlations by only matching points with consistent contour normal orientations. For each point $p_{bi}^t$ in $P_b^t$, we first conduct a nearest neighbor search to find its closest point $p_{bj}^{t-1}$ in $P_b^{t-1}$. Then we restrict our search for correlations to lines that are in the same contour as  $p_{bj}^{t-1}$. Specifically, we search for the line $l_m^{t-1} \in L_{t-1}$ that has both the shortest point-to-line distance and aligns with the contour normal vector of $p_{bi}^t$. This line, denoted as $l_m^{t-1}$, is then regarded as the point-to-line correspondence for $p_{bi}^t$.

\textbf{Point-to-line distance calculation with fitted line features.} Once the association between $p_{bi}^t$ and $l_m^{t-1}$ is established, we utilize the point-to-line distance directly between $p_{bi}^t$ and $l_m^{t-1}$ to formulate an optimization problem to mitigate potential noise that may arise from calculating the point-to-line distance using the two nearest points.

\begin{figure}[htbp]
    \centering
    \includegraphics[width=0.9\linewidth]{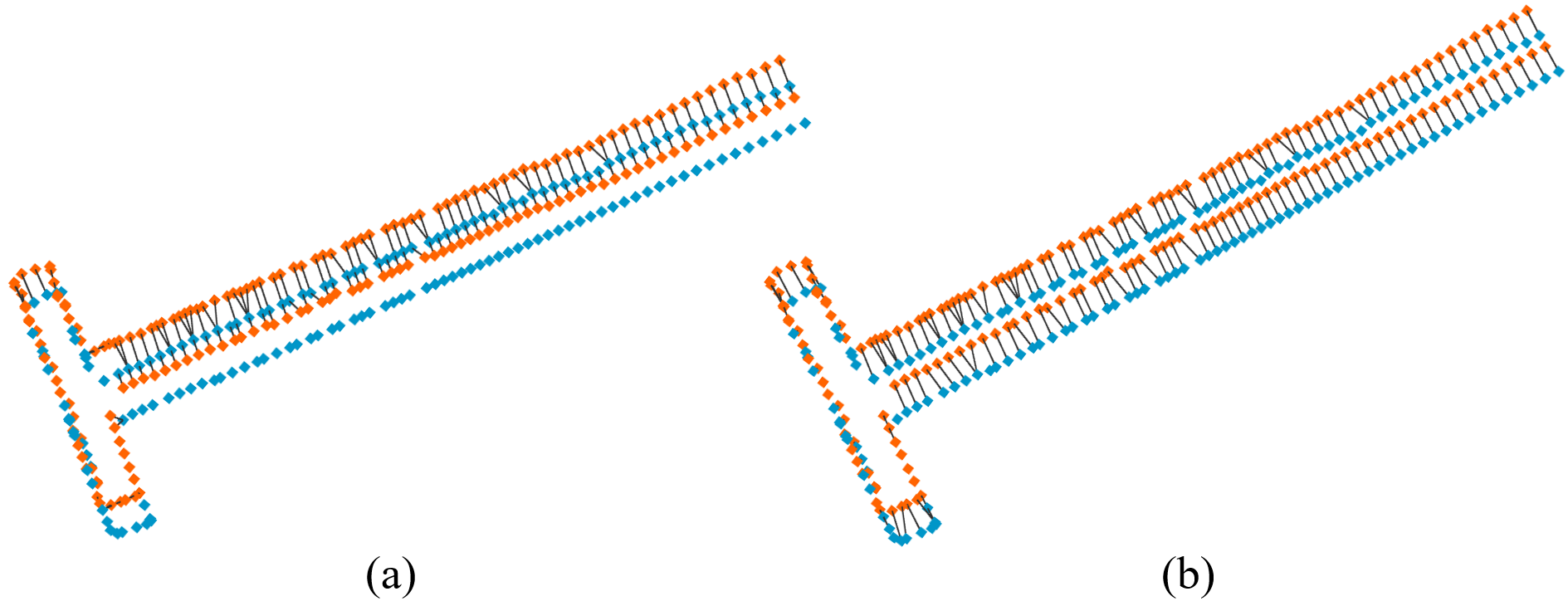}
    \caption{Matching results of a pair of 
 adjacent frames obtained by different methods. (a) The matching result formed by the nearest search. (b) The matching result formed by our method.}
    \label{fig-match}
\end{figure}

As shown in Fig. \ref{fig-match}, data association based on nearest neighbor search typically requires a lot of iterations due to inaccurate point-to-point matching. However, with the assistance of contour normal vectors, we can have more accurate data associations, achieving a comparable level of accuracy with a higher efficiency. Additionally, we leverage IMU preintegration\cite{preintergrate} to further refine the results of odometry.

\subsection{Local Mapping}
\label{localMapping}

We obtain the transformation between consecutive frames through odometry. A keyframe is generated when the robot's pose change surpasses a predefined threshold. Due to the limitation of semantic information and the existence of fragmented, low-confidence observations in individual measurements, we employ probabilistic filtering to fuse all observations between two keyframes, resulting in a better-fused keyframe.

We define each ordinary frame as $F^i=\{^{k}_{i}T,L^i P_{b}^{i}\}$, where $^{k}_{i}T$ represents the pose transformation between $F^i$ and the previous keyframe $\boldsymbol{F}_k$. We associate line features in each ordinary frame with the keyframe, merging identical line features from multiple ordinary frames into a batch denoted as $b^k_j=(l^k_j,\{l_{m_1}^{i_1},l^{i_2}_{m_2},...l^{i_n}_{m_n}\})$, where $l_{m_1}^{i_1}$ represents the $m_1$-th line feature obtained from the $i_1$-th ordinary frame. $l^k_j$ represents the fused line feature obtained from all lines in $b^k_j$. We define a keyframe as $\boldsymbol{F}_k=\{^w_kT,B^k\}$, where $^w_kT$ is the global pose corresponding to the keyframe, and $B_k=\{b^k_1,b^k_2,...,b^k_n\}$ represents all batches in the keyframe.

We assess whether a line belongs to a batch based on contour normal vectors, line directions, and line-to-line distances. Additionally, each batch $b^k$ is assigned a confidence score $p^k$ that indicates its stability. This score is incrementally updated by a Bayesian filter with the formula: 
\begin{equation}
p^k(b^k) = p^k(b^k|l_{m_1}^{i_1},..., l^{i_{n-1}}_{m_{n-1}}) + \log{\frac{p(b^k|z_{l^{i_n}_{m_n}})}{1-p(b^k|z_{l^{i_n}_{m_n}})}} 
\end{equation} 
where $p(b^k|z_{l^{i_n}_{m_n}})$ represents the confidence score of line segment $l^{i_n}_{m_n}$ obtained from the calculations in Section \ref{sectionSemantic}. We average the best-observed lines in $b^k_l$ to obtain $l^k_j$, allowing us to form a set of lines $L^k=\{l^k_1,...l^k_n\}$ for registration tasks in a keyframe. This map fusion method preserves high-quality and the most stable observations within a keyframe while maintaining the intact structure of ground semantics.

\subsection{Loop Closure Detection and Global Optimization}
\label{loopClosure}

To achieve loop closure detection, we conduct registration between keyframes. When generating a new keyframe, we align the keyframe with nearby keyframes via point-to-line registration to identify potential loop closures. However, relying solely on the convergence of registration can occasionally lead to erroneous convergence due to local optima. Therefore, after completing the registration between the current keyframe $\boldsymbol{F}_i$ and the candidate keyframe $\boldsymbol{F}_{dst}$, we implement a similar method in the local mapping procedure to evaluate the results of align. We attempt to merge $\boldsymbol{F}_i$ into $\boldsymbol{F}_{dst}$ and calculate the ratio:
\begin{equation}
\label{overlap}
p=\frac{\sum_{j=0}^N len_j\times c_j\times s}{\sum_{j=0}^N len_j\times c_j}
\end{equation}
where $len_j$ represents the length of line feature $l_j\in \boldsymbol{F}_i$, $c_i$ is the confidence of $l_i$. And $s$ is defined as:
\begin{equation}
s=
\left\{\begin{aligned}
0 &\quad\text{if $l_j\in \boldsymbol{F}_i$ can't be merged into $\boldsymbol{F}_{dst}$}\\ 
1 &\quad\text{if $l_j\in \boldsymbol{F}_i$ can be merged into $\boldsymbol{F}_{dst}$}\\
\end{aligned}\right.
\end{equation}
Here, $p$ describes the overlap ratio between $F_i$ and $F_{dst}$. If $p$ exceeds a certain threshold, the registration is considered to have converged, indicating a loop closure.

We construct a factor graph for global pose optimization, incorporating three types of factors: (1) odometry factors, (2) IMU preintegration factors, and (3) loop closure factors. We use iSAM2 \cite{isam2} for global optimization. After completing global optimization, we can merge the local map into a global map using the optimized poses.
Since the map is composed of a set of batches accumulated over time, we only need to store some vectorized descriptions of each batch for offline map storage, such as 6-DOF parameters, contour normals, endpoints, and batch confidence. This method significantly reduces map storage.

\subsection{Localization and Map Update}
\label{update}
With the completion of the mapping process, when a vehicle revisits the same environment, we implement localization and update procedures.
In the localization process, we highlight that in addition to the prior map for guidance, the proposed algorithm can accommodate scenarios where the environment may have undergone changes or the vehicle enters areas not previously covered by the existing map.

After parameterizing the ground semantic information, we run two modules in parallel: an odometry module based on frame-to-frame registration and a localization module based on the prior map. The odometry module is consistent with the one mentioned previously. In the localization module, we load the surrounding prior map based on the pre-estimated pose and implement a frame-to-map registration to obtain the vehicle's pose and position in the global coordinate system.

\begin{figure}[htbp]
    \centering
    \includegraphics[width=1.0\linewidth]{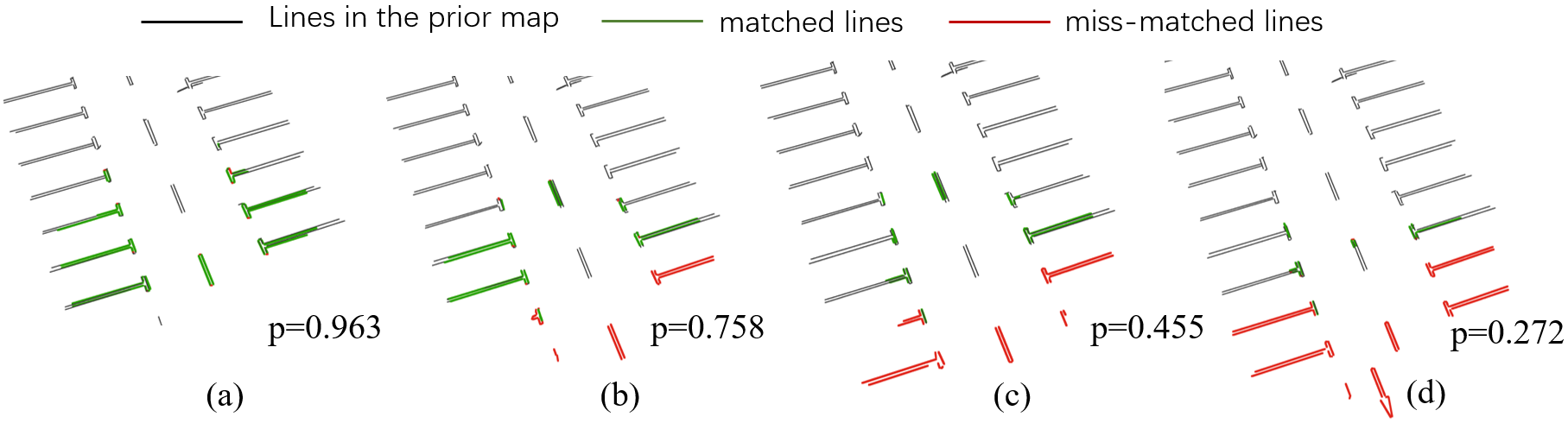}
    \caption{Illustration of the overlap value at the edge of the map. (a) to (d) represents changes of the overlap ratio during a vehicle's movement from the area covered by the map toward the map edge. (a) Fully overlapped; (b) and (c) Partially overlapped; (d) Not overlapped.}
    \label{fig-overlap}
\end{figure}

To estimate whether the vehicle has encountered environmental changes or has reached the edge of the prior map, we perform a validity assessment within the localization module. Given that the vehicle’s current observations can not establish effective correspondences with the prior map in such cases, we evaluate the overlap between the current frame and the map frame following the frame-to-map registration. The degree of overlap is quantified using (\ref{overlap}). The visualization of overlap change at the map edge is illustrated in Fig. \ref{fig-overlap}. When the calculated overlap value falls below a predefined threshold, the localization result is regarded as invalid. This validity determination is then sent to the graph optimization component.

\begin{figure}[htbp]
    \centering
    \includegraphics[width=0.9\linewidth]{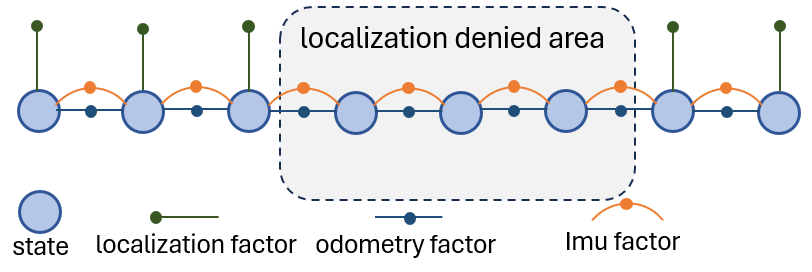}
    \caption{Illustration of the factor graph.}
    \label{fig-poseGraph}
\end{figure}

After obtaining the localization and odometry results, we maintain a factor graph as shown in Fig. \ref{fig-poseGraph}. The factor graph comprises three key types of factors: odometry factors, IMU factors, and localization factors. The odometry factors are constructed based on the frame-to-frame results obtained from the odometry module, while the localization factors are constructed based on the frame-to-map results provided by the localization module. Notably, these localization factors are only incorporated when the localization is valid.

Optimizing the factor graph can correct the accumulated drift in cases where the localization is invalid, obtaining a continuous trajectory. Since the factor graph mainly deals with trajectory alignment and optimization in unmapped areas, we only keep the most recent 10 localization results in the pose graph and treat the previous results as stable to avoid handling a large number of variables in the factor graph. 

After obtaining a continuous trajectory, we employ the method detailed in Section~\ref{localMapping} to generate local maps using current observations and the newly obtained trajectory. These local maps with global poses in keyframes, are merged into the prior map, supplementing missing parts and updating/refining the existing map.

\section{Experiments}

We validate our algorithm through real-world experiments and comparative studies. Our vehicle is equipped with four omnidirectional fisheye cameras mounted at the front, rear, and left/right side mirrors. The localization ground truth is provided by an RTK-GPS.
For comparison, we implement AVP-SLAM \cite{avpslam} and a grid-based mapping approach\cite{roadmap} that achieved ground semantic feature registration through ICP. Additionally, we use ORB-SLAM2\cite{orbslam2} as a benchmark with a forward-facing camera mounted on the roof of the vehicle to obtain sufficient features for visual odometry. We compare the performance of the proposed method with baseline methods in mapping and localization tasks. Furthermore, we conduct experiments to validate its effectiveness in tasks such as registration, map storage, and long-term localization.

\subsection{Mapping Evaluation}

We evaluate the mapping performance in an outdoor environment, using RTK-GPS as the ground truth. The vehicle runs in an outdoor parking lot, forming a trajectory with loop closures. To evaluate the mapping accuracy, we compare the absolute trajectory errors of ORB-SLAM2, the ICP-based method\cite{avpslam}, and the proposed method. The results are shown in Table \ref{mapping_evo}. Our method achieves higher mapping accuracy while requiring smaller map storage.

\begin{table}[h]
\footnotesize
\setlength\tabcolsep{3.8pt}
\caption{evaluation results of mapping trajectory error and map size.}
\vspace{-1mm}
\label{mapping_evo}
\centering
\begin{tabular}{c|ccccc}
\toprule
Method & max (cm) & RMSE (cm) & NEES (\%) & memory \\
\cmidrule{1-5}
ORB-SLAM2\cite{orbslam2} & 1.077& 0.5761& 1.679 &55.2MB\\
AVP-SLAM\cite{avpslam} & 0.724& 0.351& 0.563& 2.8MB\\
Ours & \textbf{0.561}& \textbf{0.260}& \textbf{0.511} & \textbf{433KB}\\
\bottomrule
\end{tabular}
\begin{tablenotes}    
        \footnotesize               
        \item *RMSE is the root mean square error. NEES is the normalized estimation error squared, which equals to RMSE/total length.
\end{tablenotes}
\end{table}

We compare the efficiency between our method and the ICP-based method\cite{avpslam}. In the experiments, since our method requires fewer matching and optimization steps, the time consumption of our method takes an average of only \textbf{12.25 ms}, while the conventional ICP-based methods require 60 ms to achieve a comparable level of accuracy. 




\subsection{Localization Evaluation}

To evaluate the localization accuracy and robustness of our proposed method, we compare the localization performance under different environmental conditions. Additionally, we compare the performance of our method with a conventional method\cite{roadmap} that is widely used.

When evaluating localization results, we focus on assessing the registration accuracy of frame-to-map alignment since practical tasks such as automated parking require precise localization on a local map rather than a global map. 
To quantitatively evaluate localization accuracy, we employ a two-step process. Firstly, we use a sequence of outdoor data as the mapping dataset and utilize ground truth poses to create the map. Secondly, we conduct localization on another sequence of data using the previously constructed map. Then the frame-to-map registration error can be evaluated by the trajectory error between the ground truth and the estimated trajectory.

We select a dataset captured on a sunny day for mapping and compare the localization results of our method under varying environmental conditions. Notably, the lighting conditions and the presence of parked vehicles in the parking lot during the localization sequence differ significantly from those during the mapping sequence. We use the method proposed in \cite{roadmap} as our benchmark for comparison, which involves creating a localization-oriented grid map by accumulating probability values on the grid. We compute the registration results without relying on any sensor fusion, the trajectory errors are detailed in Table \ref{table_envLoc}, where the data on the left represents the localization results based on grid map\cite{roadmap} while the data on the right represents ours. Our method demonstrates remarkable accuracy under different environmental conditions and achieves higher accuracy compared to the benchmark.


 \begin{table}
\footnotesize
\caption{Localization comparison on various environments.}
\vspace{-1mm}
\label{table_envLoc}
\centering
\begin{tabular}{c|ccc}
\toprule
Environment & Max (cm) & Avg (cm) & RMSE (cm) \\
\cmidrule{1-4}
 & \multicolumn{3}{c}{Grid-map based localization\cite{roadmap} / Ours} \\
\cmidrule{1-4}
Sunny & 0.583\;/\;\textbf{0.386} &0.156\;/\;\textbf{0.067} & 0.194\;/\;\textbf{0.095} \\
Dusk & 0.584\;/\;\textbf{0.441} &0.164\;/\;\textbf{0.071} & 0.197\;/\;\textbf{0.089} \\
Cloudy & 0.688\;/\;\textbf{0.602}& 0.192\;/\;\textbf{0.136} & 0.235\;/\;\textbf{0.143}\\
\bottomrule
\end{tabular}
\end{table}

\subsection{Long-term Map Update Evaluation}
\begin{figure}[h]
    \centering
    \includegraphics[width=1.0\linewidth]{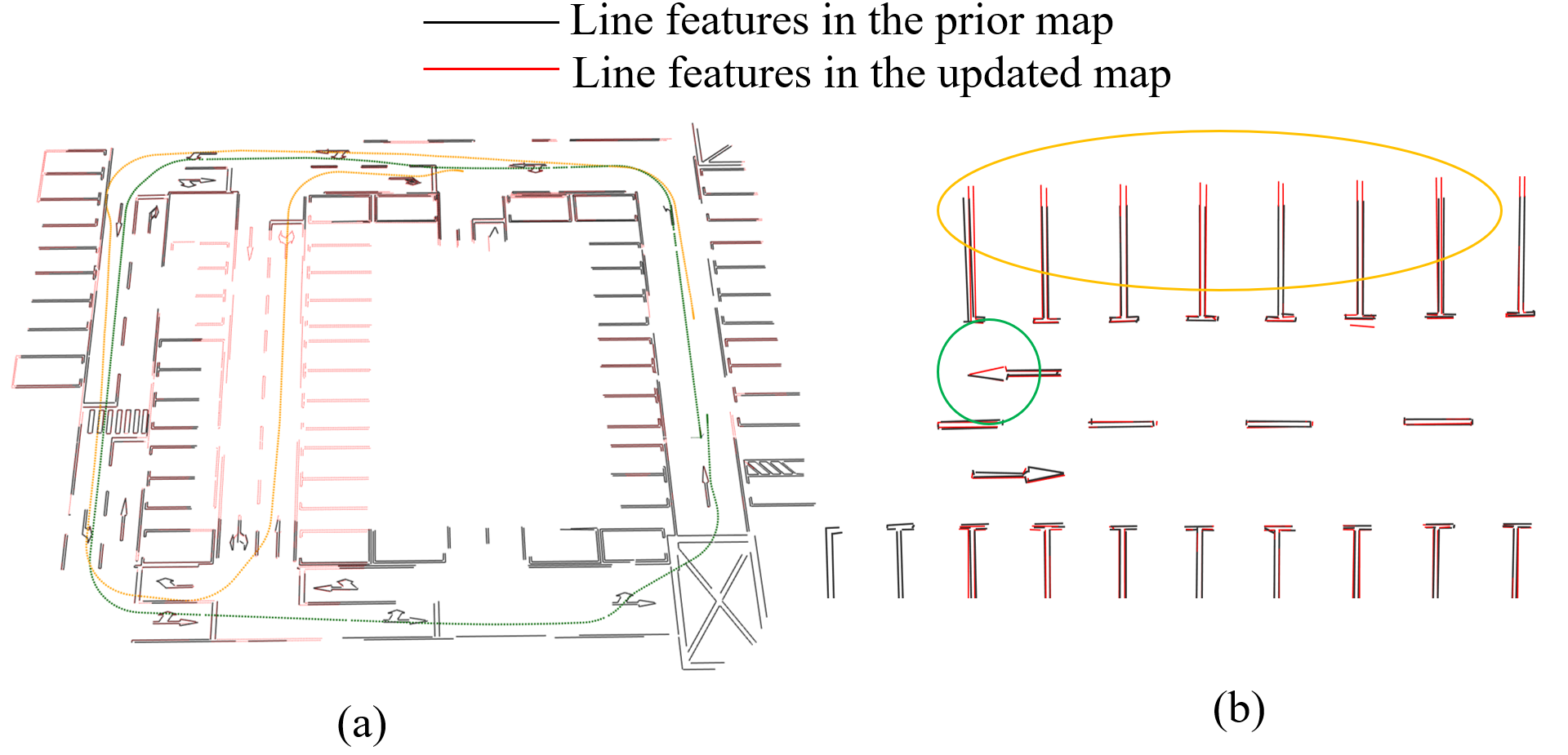}
    \caption{Map update results. (a) Map completion results in an underground parking lot, the green trajectory represents the path during map creation, while the yellow trajectory represents the path during localization. (b) Map update result, the green circle highlights an arrow that has been completed, while the yellow circle highlights updated parking lines with more confidence.}
    \label{fig-merge}
\end{figure}

\textbf{Map Completion:} During localization, when the vehicle enters an area not covered by the map, we can achieve map completion based on our method described in Section \ref{update}. The results of map completion are shown in Fig. \ref{fig-merge}(a). When the vehicle enters a new road segment, the proposed method will merge the new map into the global map.

\textbf{Map Refinement:} When the vehicle is localized within the areas covered by the map, previous mapping results with missing or low-confidence observations can be updated during the localization process. The results of map updating are shown in Fig. \ref{fig-merge}(b). We compare the performance of localization on selected map regions before and after the map refinement, these map regions have undergone noticeable update. The results, as shown in Table \ref{table_update}, demonstrate that by utilizing the updated map, the localization accuracy can be significantly improved.

\begin{table}[h]
\footnotesize
\setlength\tabcolsep{3.8pt}
\caption{Absolute Pose Error (APE) before and after Map Update}
\vspace{-1mm}
\label{table_update}
\centering
\begin{tabular}{c|ccc}
\toprule
prior map & max (cm) & avg (cm) & RMSE (cm)\\
\cmidrule{1-4}
original &0.153	&0.112	&0.125\\
updated & \textbf{0.127} & \textbf{0.075}& \textbf{0.082}\\
\bottomrule
\end{tabular}
\vspace{-3mm}
\end{table}

\section{Conclusions}

This paper presents a comprehensive solution for precise localization in parking lots by leveraging ground semantic features with low-cost cameras. We offer an accurate, efficient, and robust mapping and localization solution along with an update scheme to make the map evolve with changes in the environment. The practicality of our system has been validated through real-world experiments, highlighting its potential for widespread adoption in intelligent vehicles. The experimental results demonstrate that the proposed method outperforms state-of-the-art algorithms in terms of accuracy and robustness. In the future, we focus on generalizing the proposed system for more scenarios and extending it into a multi-agent system with higher practicality and scalability.

{\small
\bibliographystyle{ieeetr}
\bibliography{ref}
}

\end{document}